\documentclass{article}
\usepackage{PRIMEarxiv}

\usepackage{times}

\usepackage{soul}
\usepackage{url}
\usepackage{hyperref}
\usepackage[utf8]{inputenc}
\usepackage[small]{caption}
\usepackage{graphicx}
\usepackage{mathrsfs}
\usepackage{amsmath}
\usepackage{amsthm}
\usepackage{booktabs}
\usepackage{algorithm}
\usepackage{algpseudocode}
\usepackage{amssymb}
\usepackage{subfigure}
\usepackage{multirow}

\usepackage{epstopdf}
\usepackage{amsfonts}
\usepackage{color}
\usepackage{balance}
\usepackage{threeparttable}
\usepackage{abstract}

\definecolor{airforceblue}{rgb}{0.36, 0.54, 0.66}

\newcommand{\hide}[1]{}

\newtheorem{problem}{Problem}
\newtheorem{lemma}{Lemma}

\newcommand{\name}{\textsc{DetecTA}}

\usepackage{enumitem}

\title{Detecting Topology Attacks against Graph Neural Networks}

\author{
Senrong Xu$^1$\and 
Yuan Yao$^1$\and 
Liangyue Li$^2$ \and
Wei Yang$^3$\and
Feng Xu$^1$ \and
Hanghang Tong$^4$ \and
\\
$^1$State Key Laboratory for Novel Software Technology, Nanjing University, China \\
$^2$Alibaba Groups, China \\
$^3$University of Texas at Dallas, USA \\
$^4$University of Illinois Urbana-Champaign, USA\\
\\
srxu@smail.nju.edu.cn, \{y.yao, xf\}@nju.edu.cn,
wei.yang@utdallas.edu,
htong@illinois.edu
}

\begin{document}

\maketitle

\begin{abstract}
Graph neural networks (GNNs) have been widely used in many real applications, and recent studies have revealed their vulnerabilities against topology attacks. To address this issue, existing efforts have mainly been dedicated to improving the robustness of GNNs, while little attention has been paid to the detection of such attacks. In this work, we study the victim node detection problem under topology attacks against GNNs. Our approach is built upon the key observation rooted in the intrinsic message passing nature of GNNs. That is, the neighborhood of a victim node tends to have two competing group forces, pushing the node classification results towards the original label and the targeted label, respectively. Based on this observation, we propose to detect victim nodes by deliberately designing an effective measurement of the neighborhood variance for each node.  Extensive experimental results on four real-world datasets and five existing topology attacks show the effectiveness and efficiency of the proposed detection approach.
\end{abstract}


\section{Introduction}\label{sec:introduction}
Due to the strong empirical performance, GNNs have been widely studied in applications with various graph data~\cite{zhang2020deep}.
However, recent studies have found that GNNs can be easily compromised by {\em topology attacks}~\cite{zugner2018adversarial,dai2018adversarial}, i.e., inserting/deleting a few edges in the graph so as to mislead the prediction results. 
Such vulnerability significantly reduces users' confidence of using GNNs especially in security-critical applications.

To defend against topology attacks on GNNs, existing work has been mainly devoted to either pre-processing the graph data to filter out a subset of suspicious edges~\cite{wu2019adversarial,entezari2020all}, or developing novel GNNs to reduce the negative effect caused by the attacks~\cite{zhu2019robust,jin2020graph}. However, these defense techniques usually come at a cost of reducing the accuracy of GNNs.

On the other side, little attention has been paid to the {\em detection} of such attacks, e.g., identifying victim nodes under attack. Detecting such attacks is non-intrusive, and thus will not affect accuracy. Additionally, if we can identify the victim nodes beforehand, protective actions can be taken to prevent future losses.
However, detecting topology attacks is challenging due to the fact that existing attacks usually perturb only a small amount of edges, and by their design the perturbations are meant to be  {\em unnoticeable} in terms of affecting the graph's global structural properties~\cite{zugner2018adversarial}.

\hide{
\begin{figure}[t]
    \center
    \includegraphics[width=3.2in]{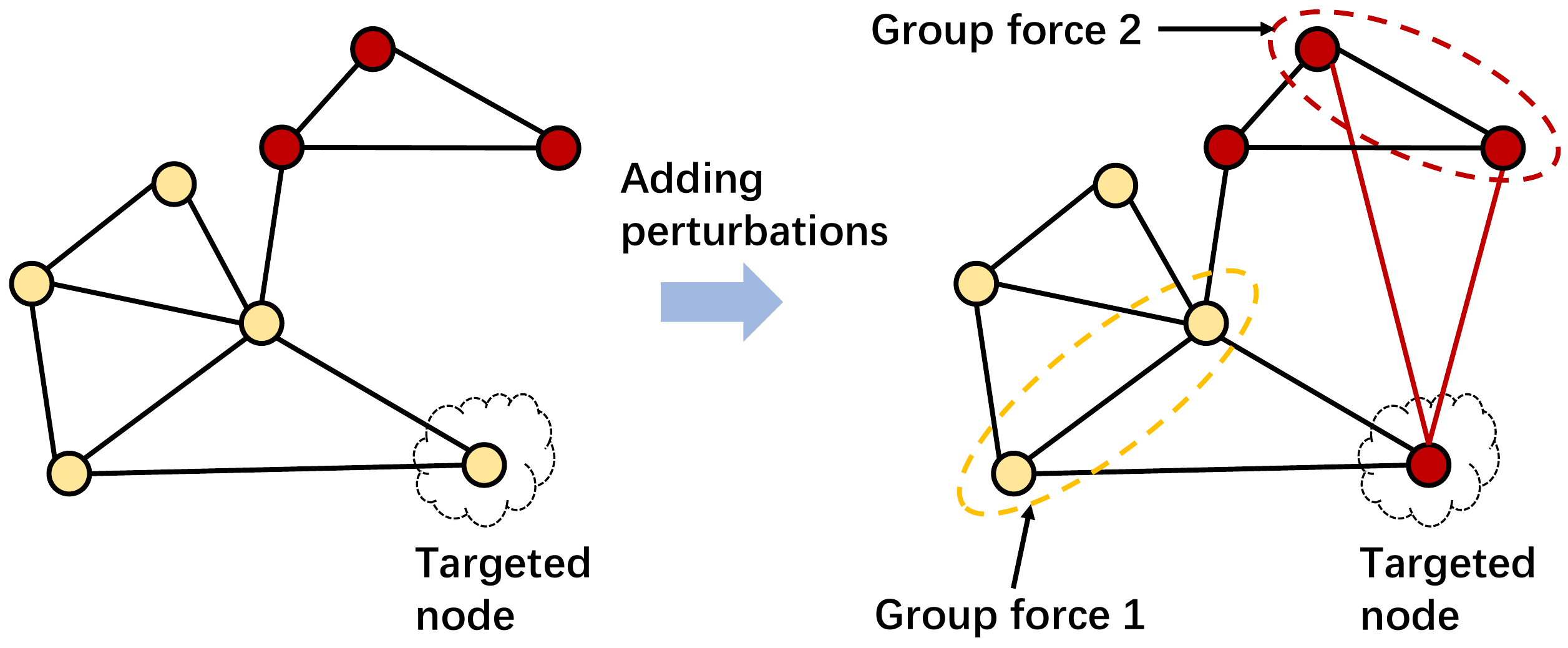}
    \caption{An illustrative example.}
    \label{F:obs}
\end{figure}
}
\hide{An example is illustrated in Fig.~\ref{F:obs}. Suppose a topology attack aims to change the classification label of the node in the lower right corner, whose original label is indicated by light yellow. This attack performs perturbations by adding two edges into the graph, which successfully change the label to the targeted one indicated by dark red. In this case, the neighborhood of the victim node has two competing group forces (i.e., group force 1 leading to the original label and group force 2 leading to the targeted label).}

In this work, we study the victim node detection problem against GNNs' topology attacks.
Our key observation is that, most GNNs are built upon the message passing framework with rooted assumption of homophily~\cite{zhu2021graph}, and the feature of each node is recursively aggregated from its neighbors; therefore, the neighborhood of a victim node would tend to have two competing group forces, pushing the node classification results towards the original label and the targeted label, respectively.
Based on this observation, we design our approach \name, \underline{detec}ting \underline{t}opology \underline{a}ttacks against GNNs. Specifically, instead of directly measuring neighborhood variance, \name\ is built upon the following two insights for a better measurement in terms of victim detection.
First, neighbor features are inherently noisy especially when the central node is under attack, making the variance computation alone less reliable. Therefore, we propose to improve the reliability of neighbor features by incorporating the local community structure to update their features. 
Second, topology attacks are usually greedy, enforcing the prediction probabilities of victim nodes’ neighbors to be concentrated on the targeted label. This greedy nature will mainly amplify the variance on the direction of the first principal component, and thus we propose to compute the neighborhood variance only on this direction.
Extensive experiments are conducted on four real datasets with five existing topology attacks (three targeted attacks and two global attacks), and the results show the effectiveness and efficiency of the proposed approach.


The main contributions of this paper include:
\begin{itemize}
    \item A new detection approach \name\ against topology attacks on GNNs, with a special design for measuring neighborhood variance. 
    \item Experimental evaluations showing the superior performance of \name. For example, it improves the existing competitors by 7.9\% - 37.7\% on average in terms of the AUC score.
\end{itemize}


\section{Background and Problem Statement}

{\bf (A) Graph Neural Networks}.
GNNs have achieved great success in graph modeling and prediction. Take GCN~\cite{kipf2016semi} as an example. Define a graph as $G = (V,E,X)$ where $V$ is the node set, $E$ is the edge set, and $X$ is the node attribute matrix. We further use $A$ to denote the adjacency matrix of $G$, and $Y$ to denote the known labels for a subset of nodes. Then, a two-layer GCN is defined as,
\small
\begin{eqnarray}\label{E:gcn}
f(G) = \text{softmax}(\hat{A}~ \sigma(\hat{A} X W^{(0)})~ W^{(1)}),
\end{eqnarray}
\normalsize
where $\sigma(\cdot)$ is an activation function (e.g., ReLU), and $W^{(i)}$ contains the trainable parameters of the $i$-th layer. $\hat{A}$ is defined as $\hat{A} = \tilde{D}^{-\frac{1}{2}}\tilde{A}\tilde{D}^{-\frac{1}{2}}$,
where $\tilde{A} = A+I$, and $\tilde{D}$ is the diagonal degree matrix of $\tilde{A}$. In each layer, each node's feature can be seen as the aggregation of features from its direct neighbors and the node self.
GNNs are usually trained in a semi-supervised manner, 
\small
\begin{eqnarray}\label{E:trainGNN}
\min_{\theta} L_{train}(G) = \sum_{y_i \in Y} g(f_{\theta}(G)_i, y_i),
\end{eqnarray}
\normalsize
where $g$ is the loss function (e.g., cross-entropy), and $f_{\theta}(G)_i$ and $y_i$ are the predicted label and true label of the $i$-th node.
 


\noindent{\bf (B) Topology Attacks}.
The goal of topology attacks against GNNs is to induce wrong predictions for nodes by perturbing the topology structure. In this work, we consider the cases of inserting and/or deleting edges. Formally, topology attacks can be formulated as the following bi-level problem,
\small
\begin{eqnarray}\label{E:attackGNN}
\max_{\Delta G} L_{attack}(G) = \sum_{v \in V_t} h(f_{\theta^*}(G+\Delta G)_v, y_v) \nonumber\\
s.t., ~~~~~ \theta^* = arg \min_{\theta} L_{train}(G+\Delta G),
\end{eqnarray}
\normalsize
where perturbation $\Delta G$ includes edge insertions/deletions, and is usually very small with $|\Delta G|$ called the perturbation budget, $h$ is the attacking utility function (e.g., measuring whether the two labels are different), and $y_v$ is the original predicted label of node $v$. 

Eq.~\eqref{E:attackGNN} is applicable to both {\em targeted attacks} and {\em global attacks}. A targeted attack aims to mislead the predictions of certain targeted nodes (i.e., $V_t \subset V$ in Eq.~\eqref{E:attackGNN}), and we define $V_t$ as the victim node set in this case. There is a special type of targeted attacks, {\em indirect attack}, which does not perturb the direct neighborhood of victim nodes. A global attack aims to degenerate the overall performance of GNNs without specific targets (i.e., $V_t = V$), and thus we define the nodes whose neighborhood is perturbed after the attack as victim nodes. 

The goal of Eq.~\eqref{E:attackGNN} is to search for $\Delta G$ with maximum attacking utility.
For example, the existing attack 
FGA~\cite{chen2018fast} directly computes the gradient over $\Delta G$ and picks the structure perturbation with the largest impact on the attacking utility until a budget is reached.




\noindent{\bf (C) Victim Node Detection}.
In this work, we aim to identify the victim nodes that are under attack, given (1) a poisoned graph $G_p=G+\Delta G$, which has been poisoned by an unknown topology attack $L_{attack}$ with an unknown perturbation $\Delta G$, and (2) a small set of known labels $Y$, which is used to train the GNN model $L_{train}$ on graph $G_p$.

\section{The Detection Approach}


We next present the proposed detection approach \name. Given a poisoned graph $G_p$, \name\ first trains the following surrogate model,
\small
\begin{equation}\label{E:linearGCN}
f'_{\theta^*}(\hat{A}_p,X) = \text{softmax}(\hat{A}_p^2XW),
\end{equation} 
\normalsize
where $A_p$ is the adjacency matrix of $G_p$, and $\hat{A}_p$ is defined as $\hat{A}$ in Eq.~\eqref{E:gcn}. The reason for training the linearized surrogate model is that this model has been shown to be an accurate surrogate of existing GNNs such as GCN~\cite{wu2019simplifying,zugner2018adversarial}. 
After that, \name\ returns the victims by computing neighborhood variance.

\hide{
\begin{figure}[t]
    \center
    \includegraphics[width=3.3in]{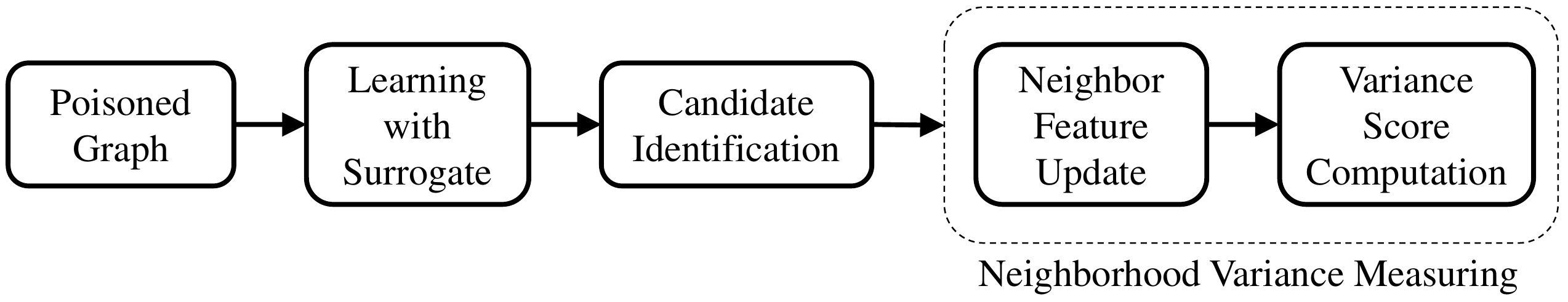}
    \caption{The overview of \name.}
    \label{F:overview}
\end{figure}
}


\subsection{Updating Neighbor Features} 

Existing work directly measures the neighborhood variance~\cite{zhang2019comparing,zhang2021detection}. However, this would be less distinguishable due to the fact that the message passing and aggregation of GNNs tend to overly smooth node features. Therefore, we propose an ``un-smoothing'' step before computing the variance. 
Specifically, for a neighbor node $u$ whose feature we aim to update, we propose a {\em biased random walk} method to compute the contribution of $u$'s neighbors to $u$'s current label prediction as follows,
\small
\begin{equation}\label{E:val}
   e(u,w) = \hat{A}_p[u][u]\hat{A}_p[u][w]Z[w][c] + 
           \hat{A}_p[u][w](\hat{A}_pZ)[w][c], \nonumber
\end{equation}
\normalsize
where $w$ is the neighbor of $u$, $c$ is the predicted label of $u$, and $Z = XW$. Here, we use $Z$ instead of $f'_{\theta^*}$ since the neighbor $w$ itself may have been attacked. In this work, we use $Z[u]$ to denote matrix $Z$'s $u$-th row and $Z[u][v]$ to denote the entry in the $u$-th row and $v$-th column.

We then normalize $e(u,w)$ to obtain the transition probability from $u$ to $w$, and start $T$ random walks of length $l$ from node $u$.
For each walk $[u, v_1,...,v_l]$, 
we update $u$'s feature as,
\small
\begin{equation}\label{E:update}
R_v[u] = Z_s[u]+\sum_{i=1}^{l}\eta^i Z_s[v_i],
\end{equation}
\normalsize
where $0<\eta<1$ is a fading factor and $Z_s[u] = \text{softmax}(Z[u])$. We compute the above equation for each of the $T$ random walks and take the average results as the final updated feature for neighbor node $u$.

\begin{figure}[t]
 \centering
 \subfigure[Before neighbor updating]{\label{F:rw3}
    \includegraphics[width=2.2in]{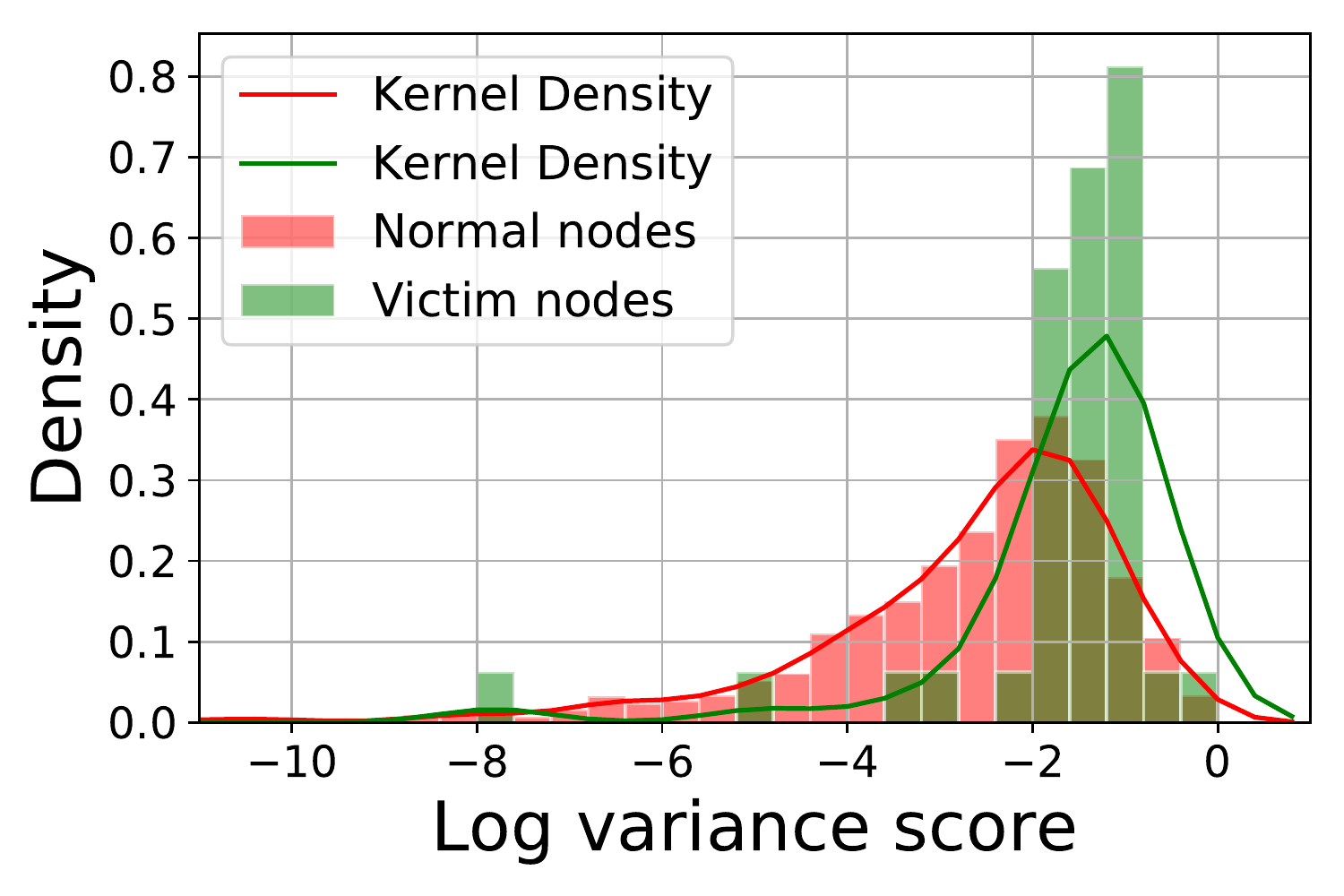}}
 \subfigure[After neighbor updating]{\label{F:rw4}
    \includegraphics[width=2.2in]{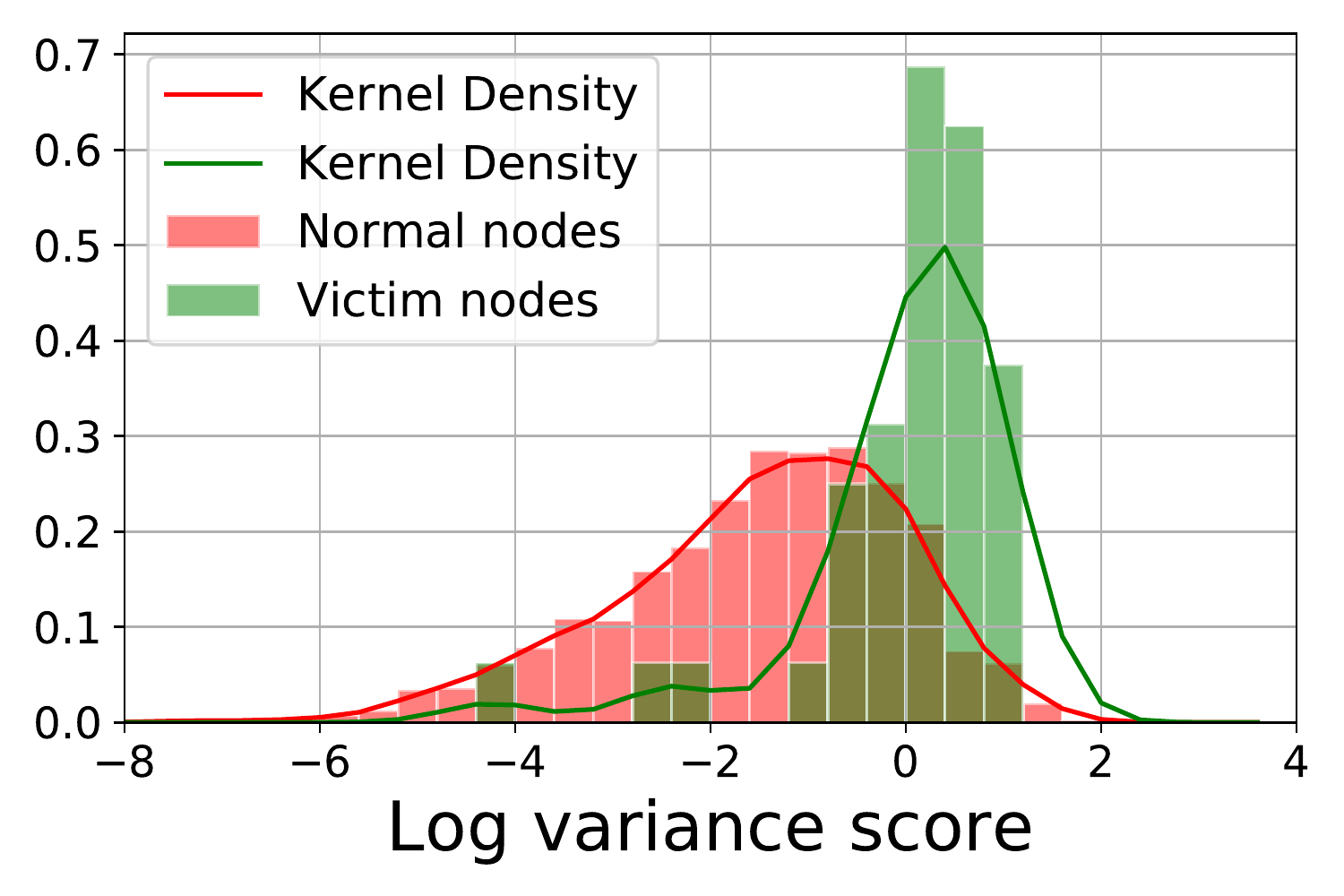}}
 \caption{The distributions of log-transformed variance scores for normal nodes (red) and victim nodes (green). Our neighbor feature updating amplifies their differences, improving AUC score from 0.83 to 0.88.}\label{F:rw_show} 
\end{figure}

{\em Remarks}. The above neighbor updating step can improve the reliability of neighbor features by incorporating their local community structure. We illustrate the effectiveness in Fig.~\ref{F:rw_show}. The two figures show the densities of neighborhood variance scores (as stated later) from normal nodes and victim nodes, before and after our neighbor updating, respectively. We can see that our neighbor updating increases the differences between normal nodes and victim nodes, and thus may lead to a better detection performance as we will later show in the experiments (e.g., AUC score improves from 0.83 to 0.88).

\subsection{Computing Variance Score}

Next, we measure the neighborhood variance for each candidate node with the updated node features.
In particular, we propose two ways to compute the variance score, i.e., {\em feature matrix} based and {\em similarity matrix} based methods.

For {feature matrix} based method, we 
directly use PCA to obtain the first principal component of $R_v$ composed of $v$'s neighbors as defined in Eq.~\eqref{E:update}, and calculate the variance of the component as the final score. This variance computation is equivalent to computing the greatest eigenvalue of the covariance matrix of $R_v$, i.e.,
\small
\begin{eqnarray}\label{E:finalscore_1}
s(v) = \lambda_1(cov(R_v)),
\end{eqnarray}
\normalsize
where $s(v)$ is the final variance score of node $v$, $cov(\cdot)$ indicates the computation of the covariance matrix, and $\lambda_i(\cdot)$ indicates the computation of the $i$-th greatest eigenvalue.

Alternatively, another choice is to emphasize the similarities between neighbors. Specifically, we define similarity matrix $W^{sim}_v$ for node $v$ based on the RBF kernel,
\small
\begin{equation}\label{E:finalscore_2}
 W^{sim}_v[i][j] = \exp{(-\kappa\lVert R_v[i]-R_v[j] \rVert^2_2)},
\end{equation}
\normalsize
where $\kappa>0$ is a kernel parameter, and each entry denotes the similarity between nodes $i$ and $j$ which are neighbors of node $v$. 
Then, we still compute the variance of the first principal component, whose computation can be further simplified as follows since $W^{sim}_v$ is a similarity matrix, 
\small
\begin{eqnarray}\label{E:finalscore_2}
s(v) 
= \max\limits_{i} (\left|\lambda_i(W^{sim}_v)\right|),
\end{eqnarray}
\normalsize
where we only need to compute the largest eigenvalue in magnitude.

\begin{figure}[t]
 \centering
    \includegraphics[width=2.8in]{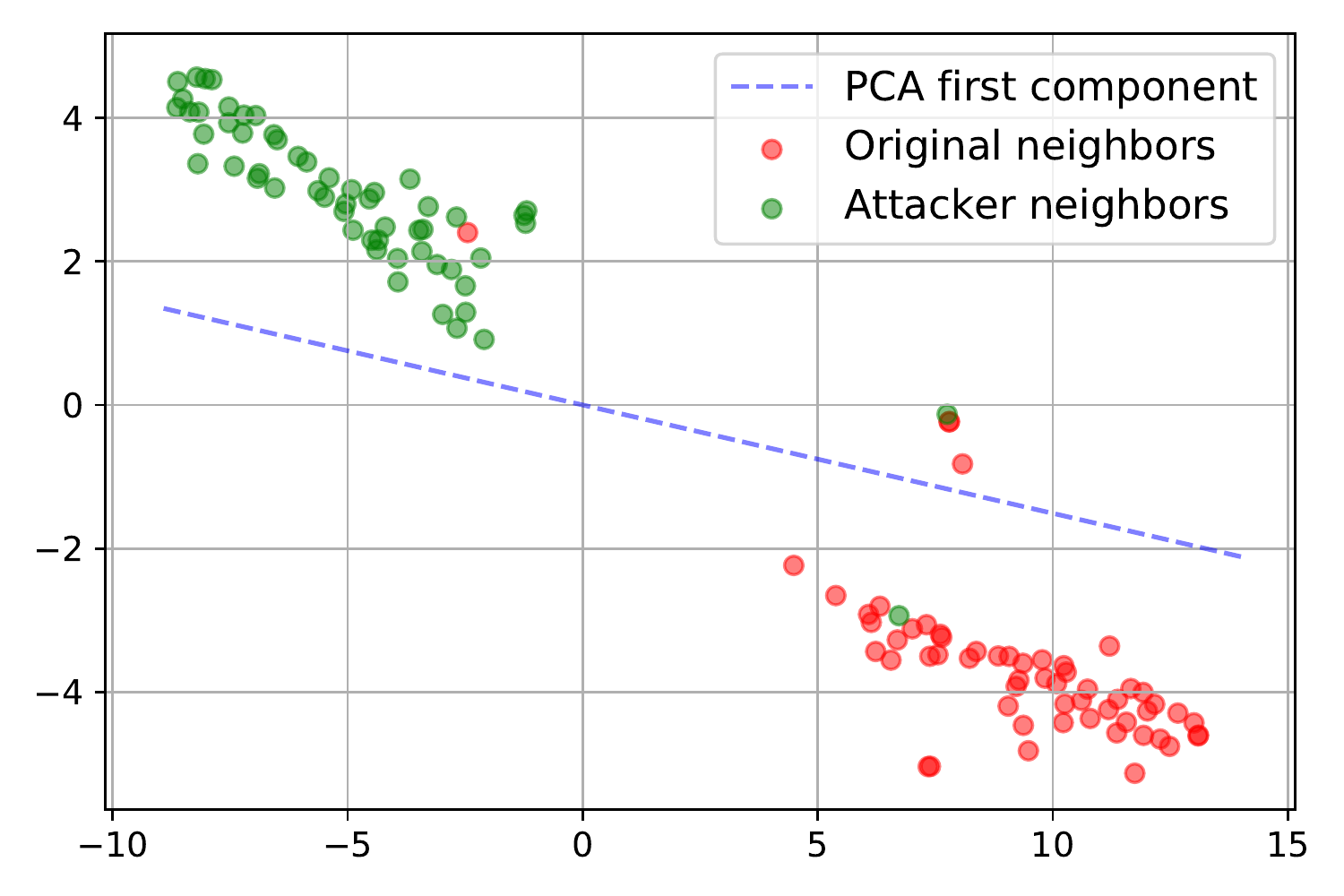}
 \caption{The intuition behind variance score computation.}\label{F:pca_show} 
\end{figure}

{\em Remarks}. Using the variance of the first principal component as our final variance score can be more distinguishable due to greedy nature of attacks. An example is shown in Fig.~\ref{F:pca_show} where the original neighbors and attacker neighbors of the victim node naturally form two groups. We can see that, the variance on the direction of the first principal component (i.e., from top left to bottom right in the figure) is most distinguishable in terms of reflecting the fact that the current node is under attack (e.g., AUC score improves from 0.83 to 0.90).

\begin{algorithm}[t]
	\caption{The \name\ Algorithm}
	\label{alg:ND}
	\begin{algorithmic}[1]
	\Require Poisoned graph $G_p=(A_p,X)$, known labels $Y$, candidate percentage $\gamma$\%, walk length $l$, walk number $T$, and fading coefficient $\eta$ 
	\Ensure The victim nodes that are under attack 
    \State train a surrogate model on $G_p$ to obtain $W$ via Eq.~\eqref{E:linearGCN};
    \State $S \gets \varnothing$;
    \State compute score $d(v)$ for each node via Eq.~\eqref{E:pcheck};
    \State $S \gets$ top $\gamma$\% nodes with highest $d(v)$ scores;
    \For{$v\in S$}
    \State initialize $R_v$;
    \For{$u\in N(v)$}
    \State $R_v[u] \gets biased$-$walk(u, l, T, \eta)$ via Eq.~\eqref{E:update};
    \EndFor
    \State compute the variance score $s(v)$ via Eq.~\eqref{E:finalscore_1}/\eqref{E:finalscore_2};
    \EndFor
    \State \Return the ranking list of nodes based on $s(v)$;
	\end{algorithmic}
\end{algorithm}

\subsection{Algorithm And Analysis}\label{sec:time}
The overall algorithm of \name\ is summarized in Alg.~\ref{alg:ND} where we assume that the underlying GNN is unknown. When the GNN is known, we can skip Line 1. In the algorithm, we first identify the candidate victim nodes (Line 3). This is an {\em optional} step.
The basic intuition is that, for a graph where homophily generally holds, a given node $v$ is probably under attack if its neighborhood's opinion is quite different from $v$'s own opinion.
We then compute the following score for each node,
\small
\begin{equation}\label{E:pcheck}
d(v) = \max [f'_{\theta^*}(\hat{A}_p,X)[v]-f'_{\theta^*}(\hat{A}'_p,X)[v]],
\end{equation}
\normalsize
where $f'_{\theta^*}(\hat{A}_p,X)[v]$ is the prediction distribution over the label space using both $v$'s own features and its neighbors' features, $f'_{\theta^*}(\hat{A}'_p,X)[v]$ is the prediction distribution only using $v$'s own features, and $A'_p$ denotes the adjacency matrix after deleting all edges connected to $v$.
Here, the score $d(v)$ indicates the maximum change of node $v$'s predicted label probability, should its neighbors exert no influence at all. 
Based on $d(v)$, we sort the nodes in the descending order and choose the top $\gamma\%$ ones for further examination (Line 4).
For each candidate node, we then update the features of its neighbors via biased random walk as indicated in Eq.~\eqref{E:update}, and then compute the neighborhood variance score (Lines 5-11).

{\em Algorithm Analysis}. The time complexity of Alg.~\ref{alg:ND} is summarized in the following lemma, which says that \name\ scales linearly w.r.t. to the size of the input graph.

\begin{lemma}\label{time analysis}
The average time complexity of Alg.~\ref{alg:ND} is $O(|V|+|E|)$ by ignoring small constants.
\end{lemma}

\hide{
Training the surrogate model in Eq.~\eqref{E:linearGCN} needs $O(|V|rC+|E|C)$ time where $r$ is the dimension of $X$ and $C$ is the number of class labels. It requires $O(|E|C)$ 
time for candidate node identification since the results $Z=XW$ can be re-used. Assuming the average node degree is $a$, then on average each candidate node needs $O(a^2TlC)$ time for updating the neighbor features, and $O(a^2)$ 
time for computing the greatest eigenvalue via the power method. 
Therefore, the overall time complexity of the algorithm is $O(|V|rC+|E|C+|V|a^2TlC)$. Since $r, C, a, T, l$ are usually small constants, the time complexity can be re-written as $O(|V|+|E|)$, which completes the proof.
\qed
}

\section{Experimental Evaluations}
\begin{table}[t]
    \centering
    \caption{Statistics of the datasets.}
      \begin{tabular}{lcccc}
      \toprule
       {Datasets} & ${\text{\#N}}_{\text{Lcc}}$ & ${\text{\#E}}_{\text{Lcc}}$ & {\#Attributes} & {\#Classes} \\
      \midrule
      Cora & 2,485 & 5,029 & 1,433 & 7 \\
      Cora-ML & 2,810 & 7,981 & 2,879 & 7 \\
      Citeseer & 2,110 & 3,668 & 3,703 & 6 \\
      Pubmed & 19,717 & 44,325 & 500 & 3 \\
      \bottomrule
      \end{tabular}
    \label{T:data}
\end{table}

\begin{table*}[t]
    \centering
    \caption{Effectiveness comparison (AUC\%) of different detection methods. \name\ generally outperforms the competitors on four datasets with five existing topology attacks. The best results are in bold and the second best are underlined.}
      \begin{tabular}{llcccccc|cc}
      \toprule
       {Datasets} & Attack & {Amen} & {Radar} & {GAE} & {Jaccard} & {Prox} & {DTA} & {\name}$_{feat}$ & {\name}$_{sim}$  \\
      \midrule
      \multirow{5}{*}{Cora} & Nettack & 77.1 & 78.5 & 52.6 & 75.4 & 80.8 & 84.9 & \underline{88.9} & {\bf 93.9} \\
                            & FGA & 72.4 & 74.5 & 57.8 & 75.1 & 83.8 & 89.5 & \underline{90.6} & {\bf 95.4} \\
                            & IG-JSMA & 84.4 & 78.6 & 76.2 & 53.5 & \underline{85.6} & 82.7 & 84.2 & {\bf 86.3} \\
                            & DICE & 56.7 & 57.3 & 68.2 & 56.2 & 64.2 & 67.1 & \underline{72.8} & {\bf 74.9} \\
                            & Mettack & 33.3 & 48.5 & 65.8 & 49.3 & 55.2 & 58.8 & {\bf 75.2} & \underline{69.9} \\\midrule
      \multirow{5}{*}{Cora-ML} & Nettack & 70.5 & 72.9 & 53.8 & 76.8 & 76.2 & 82.2 & \underline{88.0} & {\bf 89.8} \\
                            & FGA & 62.4 & 73.8 & 56.4 & 73.4 & 75.1 & 84.9 & \underline{88.2} & {\bf 91.2} \\
                            & IG-JSMA & 83.1 & 76.2 & 55.8 & 78.6 & 82.1 & 78.3 & {\bf 85.1} & \underline{84.4} \\
                            & DICE & 58.2 & 59.4 & 65.3 & 53.7 & 61.7 & 68.6 & \underline{75.1} & {\bf 78.1} \\
                            & Mettack & 22.8 & 56.6 & 67.1 & 62.6 & 72.7 & 75.9 & {\bf 93.8} & \underline{82.0} \\\midrule
      \multirow{5}{*}{Citeseer} & Nettack & 72.1 & 70.3 & 62.5 & 73.1 & 77.9 & 83.1 & \underline{88.8} & {\bf 90.0} \\
                            & FGA & 61.7 & 59.8 & 65.2 & 72.6 & 80.4 & \underline{90.3} & 88.3 & {\bf 92.5} \\
                            & IG-JSMA & 82.3 & 75.4 & 68.7 & 76.4 & 81.6 & 81.3 & \underline{82.7} & {\bf 83.2} \\
                            & DICE & 58.3 & 55.3 & 66.4 & 58.4 & 63.6 & 67.5 & \underline{74.2} & {\bf 76.2} \\
                            & Mettack & 38.9 & 48.1 & 67.3 & 49.7 & 59.0 & 70.4 & {\bf 83.9} & \underline{79.7} \\\midrule
      \multirow{5}{*}{Pubmed} & Nettack & 67.4 & 47.5 & 69.3 & 79.0 & 83.3 & 90.4 & {\bf 93.9} & \underline{93.3} \\
                            & FGA & 60.9 & 59.7 & 57.2 & 77.6 & 83.9 & 91.3 & \underline{92.1} & {\bf 92.7} \\
                            & IG-JSMA & {\bf 78.5} & 57.4 & 59.3 & 72.1 & 76.3 & 77.1 & \underline{77.5} & 76.5 \\
                            & DICE & 57.8 & 55.0 & 63.2 & 58.6 & 54.1 & 69.7 & \underline{81.2} & {\bf 82.5} \\
                            & Mettack & 35.3 & 45.2 & 60.7 & 75.9 & 58.0 & 78.4 & {\bf 91.4} & \underline{83.2} \\\midrule
      \multicolumn{2}{c}{Average Result} & 61.6 & 62.4 & 61.8 & 68.5 & 72.8 & {78.6} & {\bf 84.8} & {\bf 84.8} \\
      \bottomrule
      \end{tabular}
    \label{T:exp1}
\end{table*}


\subsection{Experimental Setup}

{\bf (A) Datasets}.
In our experiments, we use four benchmark datasets from existing work~\cite{zugner2018adversarial,chen2021understanding}, including Cora-ML~\cite{mccallum2000automating}, Cora, Citeseer, and Pubmed~\cite{sen2008collective}. All these datasets are publicly available, with statistics shown in Table~\ref{T:data}. Following the previous work~\cite{zugner2018adversarial}, we consider the largest connected component of each dataset. 

\noindent{\bf (B) Topology Attacks}. 
We apply \name\ against five existing topology attacks, including three targeted attacks \textit{Nettack}~\cite{zugner2018adversarial}, \textit{FGA}~\cite{chen2018fast}, and \textit{IG-JSMA}~\cite{wu2019adversarial}, as well as two global attacks \textit{DICE}~\cite{waniek2018hiding} and  \textit{Mettack}~\cite{zugner2019adversarial}.
In our experiments, we use GCN as the underlying model so that these attacks can be directly applied. Empirical results have also shown that the perturbations generated on GCNs are relatively universal and thus these attacks have strong transferability to other GNNs~\cite{chen2018fast}.  Nevertheless, the proposed detection approach is potentially applicable to other GNNs as long as they are built upon the message passing framework.





\noindent{\bf (C) Detection Baselines}.
For detection methods, we compare \name\ with the following baselines, and evaluate different methods using the {\em AUC} score as the metric.
\begin{itemize}
    \item {\em Amen}~\cite{perozzi2016scalable}. Amen is a graph anomaly detection method based on computing the consistency between graph structure and node attributes.
    \item \textit{Radar}~\cite{li2017radar}. It is an anomaly detection method for networks based on the residual analysis.
    \item \textit{GAE}~\cite{ding2019deep}. This method uses the reconstruction error of AutoEncoder to spot anomalies. 
    \item {\em Jaccard}~\cite{wu2019adversarial}. This method is proposed to detect attacker edges by computing the Jaccard similarity between nodes. We extend it to node detection by computing the average Jaccard similarities for a node's incident edges.
    \item \textit{Prox}~\cite{zhang2019comparing}. Prox computes the KL divergence of prediction probabilities in the local neighborhood as an indicator of being attacked.
    \item \textit{DTA}~\cite{zhang2021detection}. DTA employs statistical test to detect victim nodes based on the Jensen-Shannon Divergence (JSD) of a neighborhood.
\end{itemize}

\noindent{\bf (D) Implementations and Parameters}.
We randomly split datasets into training (10\%), validation (10\%), and testing (80\%) sets. The learning rate is set to 0.01 and dropout rate to 0.5.
For the five attacks, we follow the settings in the original papers by default. For the three targeted attacks, 40 targeted nodes are selected exactly as described in \cite{zugner2018adversarial} and the perturbation budget is $d_v$, where $d_v$ is the degree of the targeted node $v$. For the two global attacks, we set the perturbation budget to 5\% of the total number of edges in the graph.

For the compared detection methods, we follow the default settings of the original papers.
For \name, we set $\gamma = 100$ (i.e., switching off the candidate identification step) and walking length $l=3$ by default. We will study the sensitivity of these two parameters. For example, as we will later show in the experiments, we have empirically found that both effectiveness and efficiency of the proposed method can be improved when we shrink parameter $\gamma$. 
We fix $T=10$, $\eta=0.8$, and $\kappa=0.01$ as these parameters are not sensitive in a wide range. 
The proposed approach is implemented with Python, and all the experiments are run on a desktop with 6 CPU cores at 2.6G Hz.

\begin{table*}[!t]
    \centering
    \caption{Detection results of indirect targeted attack. {\name}$_{sim}$ is generally better than the other baselines.}
      \begin{tabular}{lcccccc|cc}
      \toprule
       {Datasets} & {Amen} & {Radar} & {GAE} & {Jaccard} & {Prox} & {DTA} & {\name}$_{feat}$ & {\name}$_{sim}$ \\
      \midrule
      Cora & 59.8 & 61.7 & 51.2 & 59.1 & 64.1 & \underline{64.3} & 58.7 & {\bf 66.2} \\
      Cora-ML & 58.4 & 59.2 & 52.3 & 57.6 & 54.6 & \underline{61.5} & 58.9 & {\bf 66.1} \\
      Citeseer & 48.6 & 45.4 & 58.6 & 53.0 & 60.0 & 68.4 & \underline{70.5} & {\bf 72.1} \\
      Pubmed & 55.7 & 48.5 & 55.8 & 51.9 & 54.7 & {\bf 66.9} & 58.1 & \underline{65.1} \\
      \bottomrule
      \end{tabular}
    \label{T:indirect}
\end{table*}

\subsection{Experimental Results}\label{sec:results}

{\em (A) Effectiveness comparisons}.
We first compare the overall effectiveness of different detection methods, and the results are shown in Table~\ref{T:exp1}. \name$_{feat}$ means we use the feature matrix to compute variance  (Eq.~\eqref{E:finalscore_1}), and \name$_{sim}$ means we use the similarity matrix instead (Eq.~\eqref{E:finalscore_2}).

First of all, we can observe that the proposed detection \name\ generally outperforms its competitors in all the four datasets under all the five topology attacks. On average, both \name$_{feat}$ and \name$_{sim}$ improve the existing competitors by 7.9\% - 37.7\% in terms of AUC scores. 
The first three baselines (Amen, Radar, and GAE) perform relatively poor. The reason is that these three methods are general graph anomaly detection methods, and are not specially designed to detect against topology attacks on GNNs. 
The latter three baselines (Jaccard, Prox, and DTA) are all designed for detecting topology attacks against GNNs. However, these methods directly measure the neighborhood variance by computing the Jaccard similarity, KL divergence, and JSD of the neighborhood, respectively. 
This result indicates the superiority of the proposed variance measurement.

\hide{
Second, all the detection methods perform relatively well in targeted attacks compared to global attacks. This is consistent with the intuition since targeted attacks usually add perturbations that are more focused on the victim nodes.
As for the two variants of \name, \name$_{feat}$ achieves average AUC scores 87.4\% and 80.9\%, and \name$_{sim}$ achieves average AUC scores 89.1\% and 78.3\%, under targeted attacks and global attacks, respectively. This shows that \name$_{feat}$ has a slight advantage of detecting global attacks compared to \name$_{sim}$. 
}

Next, we evaluate the detection methods under indirect attacks. This type of attack is more difficult to detect as it only adds perturbations that are two-hop or even further away from the victim nodes. For this experiment, we evaluate the indirect version of {Nettack}~\cite{zugner2018adversarial}, and the detection results are shown in Table~\ref{T:indirect}. We can observe that our detection approach {\name}$_{sim}$ generally outperforms other baselines, which demonstrates the advantage of {\name}$_{sim}$ in terms of detecting indirect targeted attacks. 

Overall, \name$_{sim}$ is better for targeted attacks (both direct and indirect), and \name$_{feat}$ is better for global attacks. The possible reason is as follows.
The kernel-based method \name$_{sim}$ performs better when the number of neighbors for victim nodes is relatively large, because such case allows \name$_{sim}$ to use feature spaces of higher dimensionality. 
Existing targeted attacks tend to choose a significant subset of high-degree targeted nodes (which are more difficult to attack), and thus \name$_{sim}$ is slightly better in this case. 
On the contrary, \name$_{feat}$ becomes better when the degree of victim nodes is relatively small, which is usually the case of global attacks, as they tend to attack low-degree nodes which are easier to manipulate.

\begin{figure}[t]
 \centering
 \subfigure[AUC vs. $\gamma$\%]{\label{F:para1}
    \includegraphics[width=2.0in]{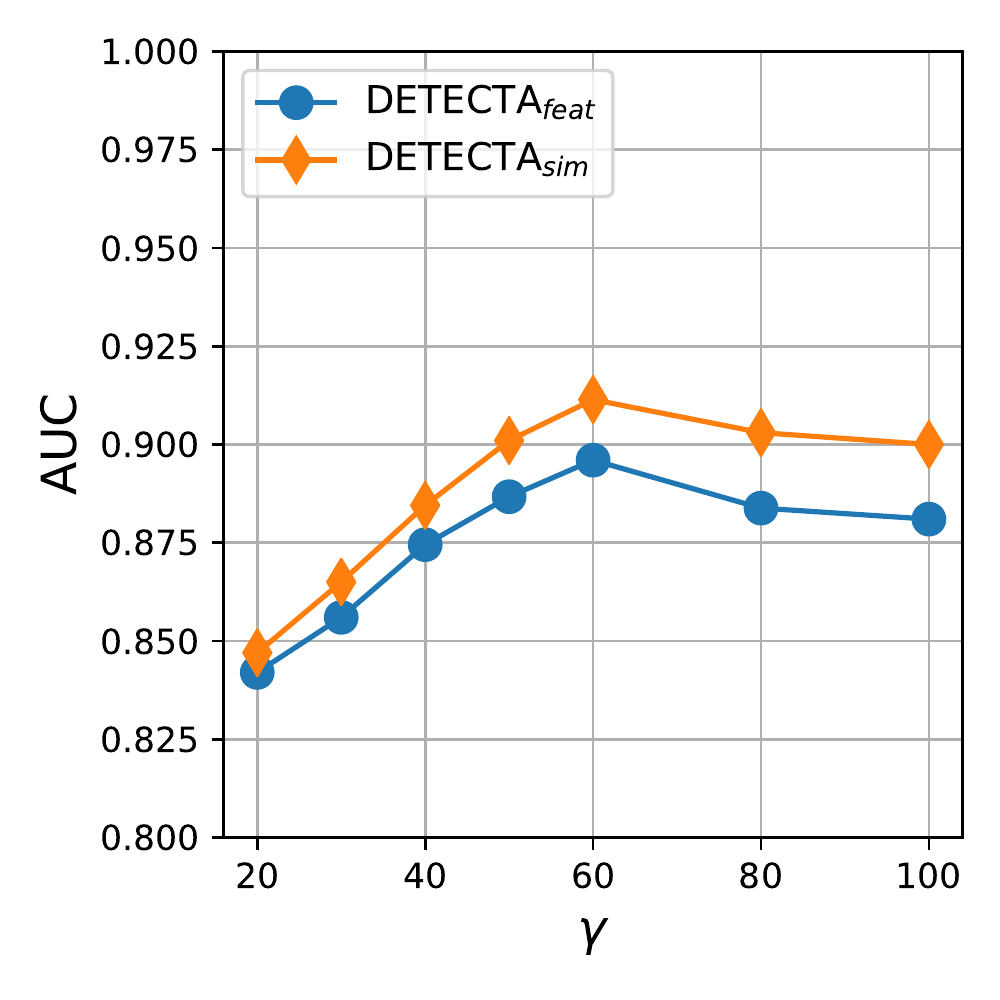}}
 \subfigure[Wall-clock time vs. $\gamma$\%]{\label{F:para2}
    \includegraphics[width=2.0in]{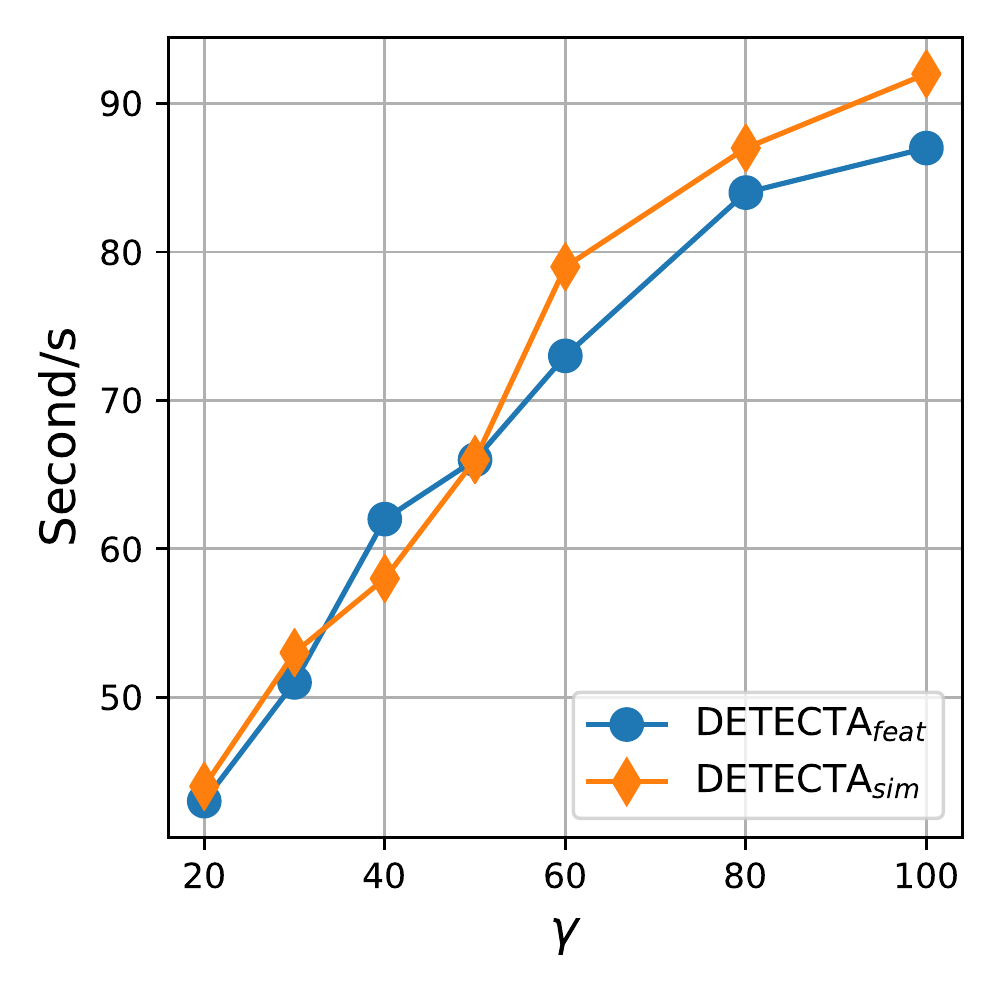}}
 \caption{The parameter sensitivity of $\gamma$. The AUC score further rises as $\gamma$ decreases from 100, and the wall-clock time scales linearly w.r.t. $\gamma$.}\label{F:para_exp1} 
\end{figure}

\begin{figure}[t]
 \centering
 \subfigure[AUC vs. $l$]{\label{F:para3}
    \includegraphics[width=2.0in]{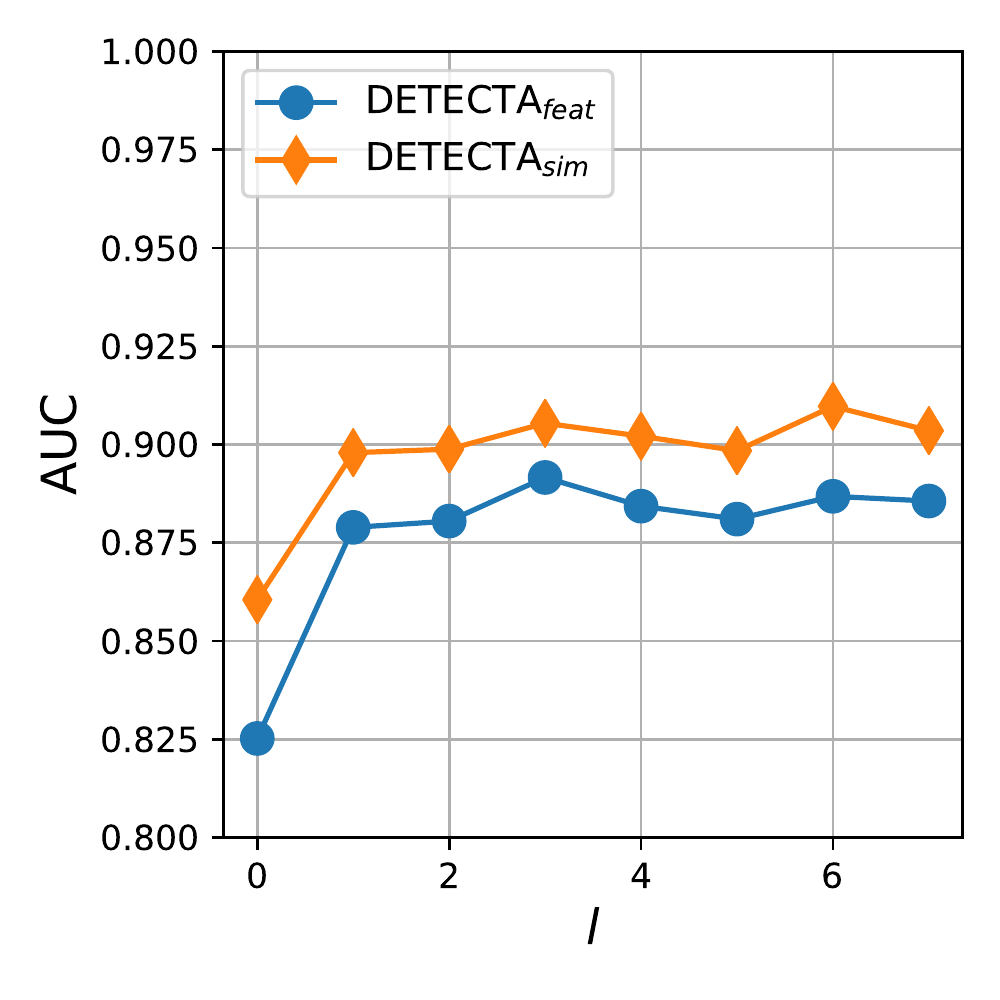}}
 \subfigure[Wall-clock time vs. $l$]{\label{F:para4}
    \includegraphics[width=2.0in]{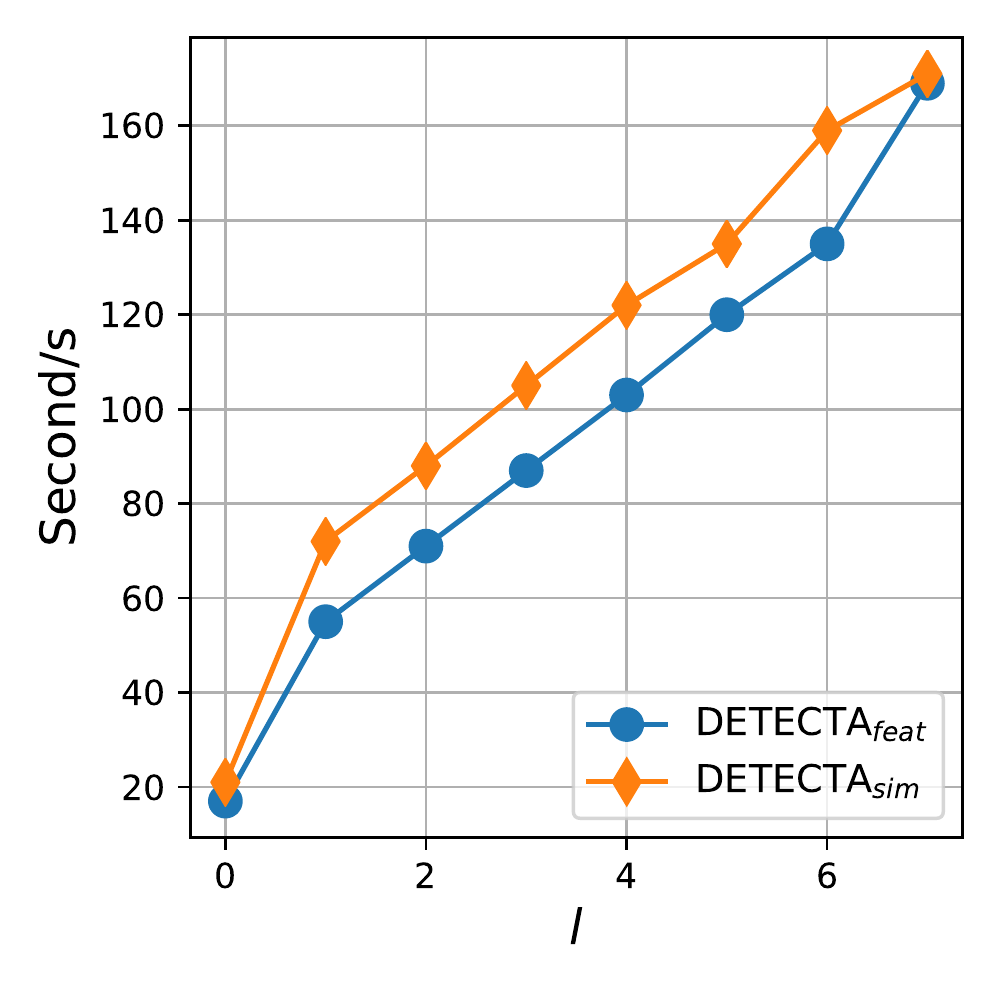}}
 \caption{The parameter sensitivity of $l$. Our neighbor updating method significantly improves the detection accuracy. Note that $l=0$ means not using neighbor update.}\label{F:para_exp2} 
\end{figure}

{\em (B) Parameter sensitivity}.
Next, we analyze the sensitivity of two parameters in \name, i.e., $\gamma$ in identifying candidate victims and walking length $l$ in updating neighbor features. For the following experiments, we show the results of Nettack on Citeseer for brevity, and similar results are observed on the other cases.
For parameter $\gamma$, we show the results in Fig.~\ref{F:para_exp1}. We can observe that the detection effectiveness of \name\ can be improved when we shrink this parameter from 100 (the default case) in a wide range. 
For example, when $\gamma=60$, the improvement of \name$_{feat}$ compared to the best competitor DTA increases by 9.6\%. 
Additionally, the wall-clock time scales linearly w.r.t. $\gamma$, which is consistent with our algorithm analysis in Section~\ref{sec:time}.
This result means that although we set $\gamma=100$ by default, both effectiveness and efficiency of \name\ can be improved when we shrink this parameter in a wide range.


For walking length $l$, we show the results in Fig.~\ref{F:para_exp2}. We can first observe from Fig.~\ref{F:para3} that our neighbor updating method significantly improves the detection accuracy. For example, the AUC score improves from 0.825 to 0.881 for \name$_{feat}$ when walking length increases from  0 to 2. Note the $l=0$ means not using neighbor update. 
Fig.~\ref{F:para4} shows that \name\ scales linearly w.r.t. $l$. 

\begin{table}[t]
    \centering
    \caption{Ablation study results. Both our neighbor updating and variance computation are better than existing methods.}
      \begin{tabular}{lcccccc|cc}
      \toprule
       {AUC} & {Nettack} & {FGA} & {IG-JSMA} & {DICE} & {Mettack}\\
      \midrule
      \name$_{feat}$ & \underline{88.8} & 88.3 & 82.6 & \underline{74.2} & {\bf 83.8} \\
      \name$_{sim}$ & {\bf 90.0} & {\bf 92.5} & {\bf 83.2} & {\bf 76.2} & \underline{79.7}  \\
      \midrule
      \name$_{feat.agg}$ & 81.2 & 75.8 & 80.9 & 72.3 & 77.6  \\
      \name$_{sim.agg}$ & 83.4 & 80.8 & 82.7 & 73.8 & 72.5  \\
      {DTA+} & 87.4 & \underline{90.6} & \underline{82.8} & 62.9 & 58.9  \\
      {Prox+} & 83.2 & 84.1 & 78.9 & 70.2 & 79.5  \\
      \bottomrule
      \end{tabular}
    \label{T:ablaion}
\end{table}

{\em (C) Ablation study}.
We next analyze the performance of different components (i.e., neighbor updating and variance computation) in measuring neighborhood variance. For updating neighbor features, we replace biased random walk with aggregation which is similar to the two-layer GCN model, i.e., $\hat{A}^2XW$. We use \name$_{feat.agg}$ and \name$_{sim.agg}$ to denote this aggregation version. For variance computation, we apply the two best existing competitors {DTA} and {Prox} upon our previous steps (denote as {DTA+} and {Prox+}). 
The results are shown in Table~\ref{T:ablaion}. We can observe that \name\ is better than \name$_{feat.agg}$, \name$_{sim.agg}$, indicating the usefulness of our neighbor updating method. 
Also, \name\ is better than {DTA+} and {Prox+} (e.g., the AUC score of \name$_{sim}$ improves {Prox+} from 0.832 to 0.900 on Nettack), indicating the usefulness of our variance computation method.


\begin{figure}[t]
 \centering
 \subfigure[Targeted attack Nettack]{\label{F:pb1}
    \includegraphics[width=2.3in]{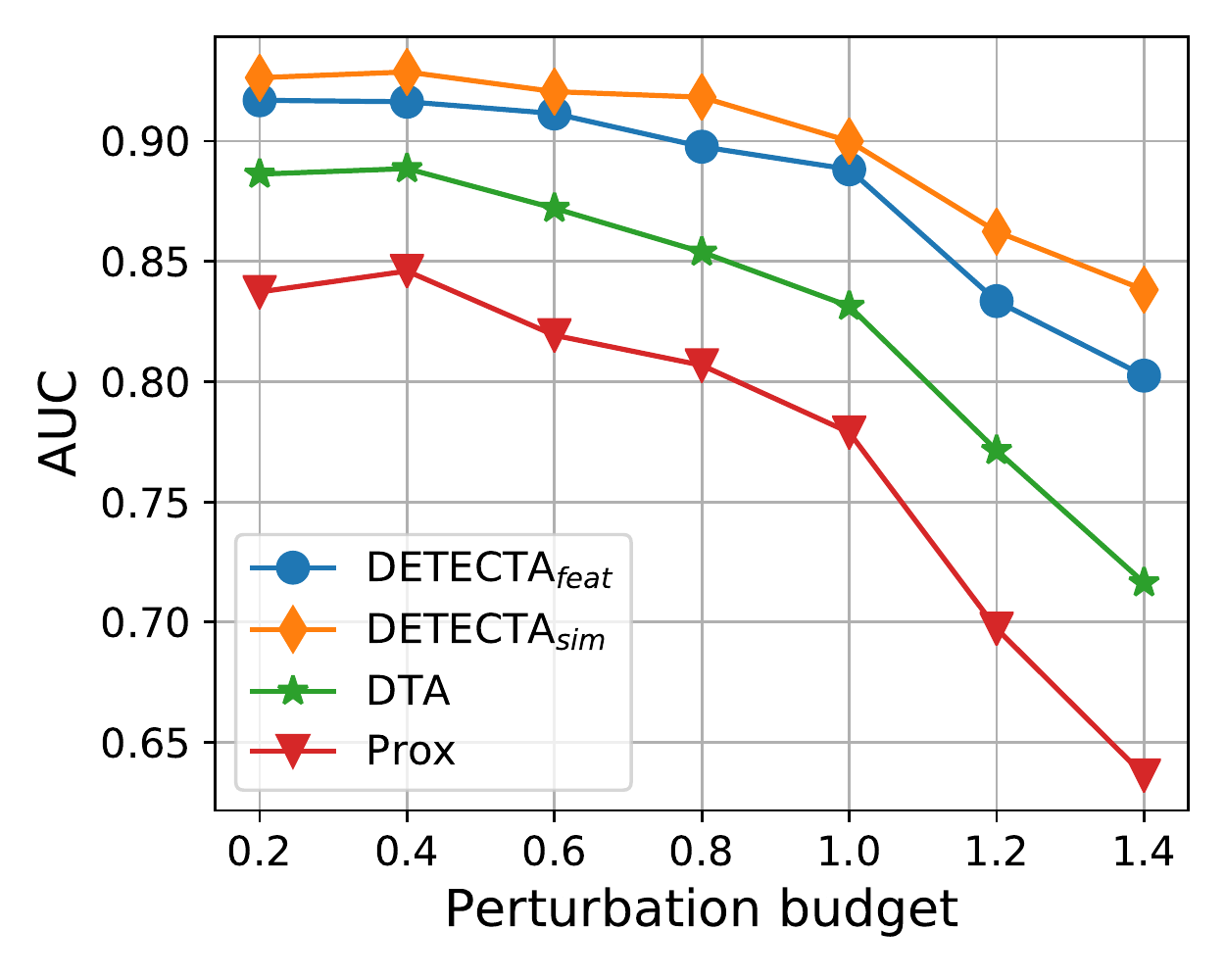}}
 \subfigure[Global attack Mettack]{\label{F:pb2}
    \includegraphics[width=2.3in]{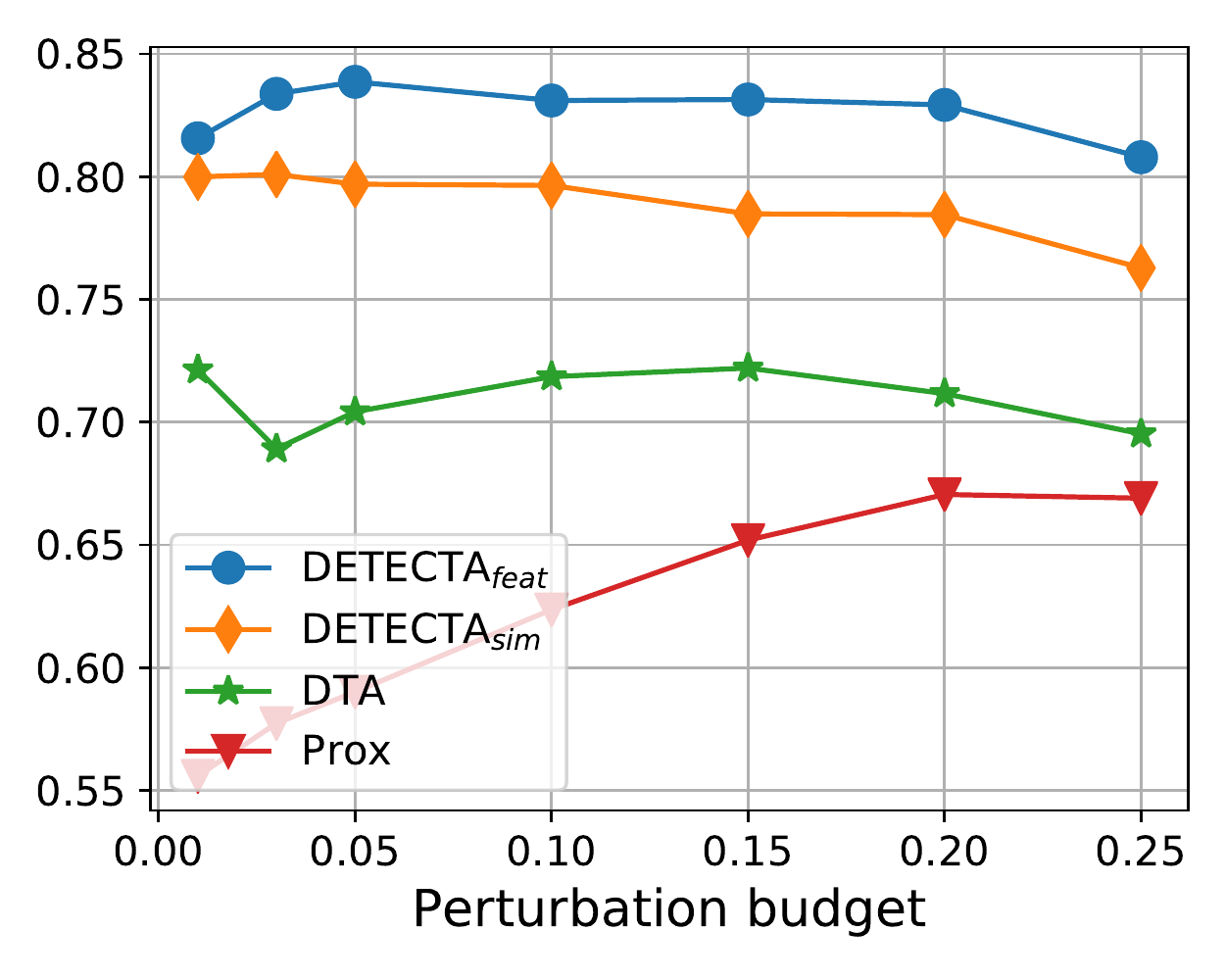}}
 \caption{Our \name\ is still better than the best competitors when the perturbation budget varies.}\label{F:pb_exp} 
\end{figure}

{\em (D) Detection results w.r.t. the perturbation budget}.
Finally, we evaluate the effectiveness of our detection method under different perturbation budgets. For targeted attack, we vary the budget as a fraction of a targeted node's degree; for global attack, we vary the overall perturbation rate. We show the results of two best competitors ({DTA} and {Prox}) against {Nettack} and {Mettack} on Citeseer in Fig.~\ref{F:pb_exp}.
We can observe that our \name\ is still better than the best competitors when the perturbation budget varies.

\section{Related Work}

{\bf GNN Attacks}.
GNN topology attacks can be divided into two classes: {\em targeted attacks}~\cite{chen2018fast,zugner2018adversarial,dai2018adversarial,wu2019adversarial,chang2020restricted,entezari2020all} and {\em global attacks}~\cite{waniek2018hiding,zugner2019adversarial,xu2019topology,ma2021graph}.
For the former, Z{\"u}gner et al.~\cite{zugner2018adversarial} propose a unnoticeable targeted attack, and Dai et al.~\cite{dai2018adversarial} study black-box targeted attacks via reinforcement learning.
For the latter, DICE~\cite{waniek2018hiding} adds/deletes edges to disturb the community structure; Mettack~\cite{zugner2019adversarial} applies meta learning and treats the graph as a hyperparameter to optimize. 
In addition to the above topology attacks, other attacks such as feature variation attacks~\cite{zugner2018adversarial} and node injection attacks~\cite{sun2020adversarial,zou21tdgia} have also been studied.

\noindent{\bf GNN Defenses}.
The first line of existing defenses apply pre-processing steps to remove suspicious or noisy edges before training the GNNs~\cite{wu2019adversarial,entezari2020all}. 
The second line directly develops new and robust GNNs to defend against attacks~\cite{zhu2019robust,zhang2020gnnguard,jin2020graph,liu2021elastic,chen2021understanding}. For example, Zhu et al.~\cite{zhu2019robust} use Gaussian distributions as the hidden node representations so that attacking effect can be absorbed; Jin et al.~\cite{jin2020graph} jointly learn the graph structure and the graph neural networks.
Other related defenses include adversarial training~\cite{xu2019topology} and measuring the safety region of each node under adversarial perturbations~\cite{bojchevski2019certifiable,wang2021certified}.

Different from the above work, we aim to detect the victim nodes that are under topology attacks. 
Currently, very few efforts have been devoted to this task with two exceptions from Zhang et al.~\cite{zhang2019comparing} and Zhang et al.~\cite{zhang2021detection}. The former computes the average KL divergences of softmax probabilities in the local neighborhood, and the latter develops a statistical test based on maximum mean discrepancy. In contrast, we deliberate a more effective way by updating neighbor features and then computing the variance of the first principal component of them.





\section{Conclusions}
In this paper, we propose \name, a novel approach to detecting victim nodes under topology attacks against GNNs. By exploiting the message passing nature of GNNs, \name\ poses a special measurement of neighborhood variance for each node as an indicator of being attacked.  Experimental results demonstrate that the proposed approach can effectively identify victim nodes from a poisoned graph. 
Future work includes extending the detection to feature variation attacks, and leveraging the detection results to better defend GNNs against adversarial attacks.


\bibliographystyle{unsrt}  
\bibliography{references.bib}

\end{document}